\documentclass{article}

 \usepackage[nonatbib,final]{nips_2016}
\usepackage{cite}

\usepackage[T1]{fontenc}    
\usepackage{booktabs}       
\usepackage{amsfonts}       
\usepackage{nicefrac}       
\usepackage{microtype}      
\usepackage{epsfig}
\usepackage{graphicx}
\usepackage{amsmath}
\usepackage{amssymb}
\usepackage{epstopdf}
\usepackage{color,soul}

\title{Evolutionary Synthesis of Deep Neural Networks via Synaptic Cluster-driven Genetic Encoding}

\author{
  Mohammad Javad Shafiee,  Alexander Wong \\
	Department of Systems Design Engineering, University of Waterloo\\
	 Waterloo, Ontario, Canada\\
	\texttt{\{mjshafiee, a28wong\}@uwaterloo.ca} \\
}
\begin{document}


\maketitle

\begin{abstract}
There has been significant recent interest towards achieving highly efficient deep neural network architectures.  A promising paradigm for achieving this is the concept of \textit{evolutionary deep intelligence}, which attempts to mimic biological evolution processes to synthesize highly-efficient deep neural networks over successive generations.  An important aspect of evolutionary deep intelligence is the genetic encoding scheme used to mimic heredity, which can have a significant impact on the quality of offspring deep neural networks. Motivated by the neurobiological phenomenon of synaptic clustering, we introduce a new genetic encoding scheme where synaptic probability is driven towards the formation of a highly sparse set of synaptic clusters.  Experimental results for the task of image classification demonstrated that the synthesized offspring networks using this synaptic cluster-driven genetic encoding scheme can achieve state-of-the-art performance while having network architectures that are not only significantly more efficient (with a $\sim$125-fold decrease in synapses for MNIST) compared to the original ancestor network, but also tailored for GPU-accelerated machine learning applications.
\end{abstract}



\section{Introduction}
\vspace{-0.4 cm}

There has been a strong interest toward obtaining highly efficient deep neural network architectures that maintain strong modeling power for different applications such as self-driving cars and smartphone applications where the available computing resources  are practically limited to a combination of low-power, embedded GPUs and CPUs with limited memory and computing power.  The optimal brain damage method~\cite{lecun1989optimal} was one of the first approaches in this area, where synapses were pruned based on their strengths.  \mbox{Gong~{\it et al.}}~\cite{gong2014compressing} proposed a network compression framework where vector quantization was leveraged to shrink the storage requirements of deep neural networks.  Han {\it et al.}~\cite{han2015learning} utilized pruning, quantization and Huffman coding to further reduce the storage requirements of deep neural networks. Hashing is another trick utilized by Chen {\it et al.}~\cite{chen2015compressing} to compress the network into a smaller amount of storage space. Low rank approximation~\cite{ioannou2015training,jaderberg2014speeding} and sparsity learning~\cite{feng2015learning,liu2015sparse,SSL_2016} are other strategies used to sparsify deep neural networks. 

Recently, Shafiee {\it et al.}~\cite{EvoNet1} tackled this problem in a very different manner by proposing a novel framework for synthesizing highly efficient deep neural networks via the idea of evolutionary synthesis.  Differing significantly from past attempts at leveraging evolutionary computing methods such as genetic algorithms for creating neural networks~\cite{white1993gannet,stanley2002evolving}, which attempted to create neural networks with high modeling capabilities in a direct but highly computationally expensive manner, the proposed novel \textit{evolutionary deep intelligence} approach mimics biological evolution mechanisms such as random mutation, natural selection, and heredity to synthesize successive generations of deep neural networks with progressively more efficient network architectures.  The architectural traits of ancestor deep neural networks are encoded via probabilistic `DNA' sequences, with new offspring networks possessing diverse network architectures synthesized stochastically based on the `DNA' from the ancestor networks and computational environmental factor models, thus mimicking random mutation, heredity, and natural selection.  These offspring networks are then trained, much like one would train a newborn, and have more efficient, more diverse network architectures while achieving powerful modeling capabilities.

An important aspect of evolutionary deep intelligence that is particular interesting and worth deeper investigation is the genetic encoding scheme used to mimic heredity, which can have a significant impact on the way architectural traits are passed down from generation to generation and thus impact the quality of descendant deep neural networks.  A more effective genetic encoding scheme can facilitate for better transfer of important genetic information from ancestor networks to allow for the synthesis of even more efficient and powerful deep neural networks in the next generation.
 As such, a deeper investigation and exploration into the incorporation of synaptic clustering into the genetic encoding scheme can be potentially fruitful for synthesizing highly efficient deep neural networks that are more geared for improving not only memory and storage requirements, but also be tailored for devices designed for highly parallel computations such as embedded GPUs.

In this study, we introduce a new synaptic cluster-driven genetic encoding scheme for synthesizing highly efficient deep neural networks over successive generations.  This is achieved through the introduction of a multi-factor synapse probability model where the synaptic probability is a product of both the probability of synthesis of a particular cluster of synapses and the probability of synthesis of a particular synapse within the synapse cluster.  This genetic encoding scheme effectively promotes the formation of synaptic clusters over successive generations while also promoting the formation of highly efficient deep neural networks.

\vspace{-0.4 cm}

\section{Methodology }
\vspace{-0.4 cm}

The proposed genetic encoding scheme decomposes synaptic probability into a multi-factor probability model, where the architectural traits of a deep neural network are encoded probabilistically as a product of the probability of synthesis of a particular cluster of synapses and the probability of synthesis of a particular synapse within the synapse cluster.

\textbf{Cluster-driven Genetic Encoding.}
Let the network architecture of a deep neural network be expressed by $\mathcal{H}(N,S)$, with $N$ denoting the set of possible neurons and $S$ denoting the set of possible synapses in the network. Each neuron $n_i \in N$ is connected via a set of synapses $\bar{s} \subset S$ to neuron $n_j \in N$ such that the synaptic connectivity $s_i \in S$ is associated with a $w_i \in W$ denoting its strength.  The architectural traits of a deep neural network in generation $g$ can be encoded by a conditional probability given its architecture at the previous generation $g-1$, denoted by $P(\mathcal{H}_g| \mathcal{H}_{g-1})$, which can be treated as the probabilistic `DNA' sequence of a deep neural network.

Without loss of generality, based on the assumption that synaptic connectivity characteristics in an ancestor network are desirable traits to be inherited by descendant networks, one can instead encode the genetic information of a deep neural network by synaptic probability $P(S_g|W_{g-1})$, where $w_{k,g-1} \in W_{g-1}$ encodes the synaptic strength of each synapse $s_{k,g} \in S_{g}$.  In the proposed genetic encoding scheme, we wish to take into consideration and incorporate the neurobiological phenomenon of synaptic clustering~\cite{welzel2010synapse, kastellakis2015synaptic, larkum2008synaptic, takahashi2012locally, winnubst2015synaptic}, where the probability of synaptic co-activation increases for correlated synapses encoding similar information that are close together on the same dendrite.

To explore the idea of promoting the formation of synaptic clusters over successive generations while also promoting the formation of highly efficient deep neural networks, the following multi-factor synaptic probability model is introduced:
\vspace{- 0.25 cm}
\begin{align}
P(S_g|W_{g-1}) = \prod_{c \in C} \Big[P\big(\bar{s}_{g,c}|W_{g-1}\big) \cdot \prod_{i \in c }P(s_{g,i}|w_{g-1,i}) \Big]
\end{align}
where the first factor (first conditional probability) models the probability of the synthesis of a particular cluster of synapses, $\bar{s}_{g,c}$, while the second factor models the probability of a particular synapse, $s_{g,i}$, within synaptic cluster $c$.   More specifically, the probability $P(\bar{s}_{g,c}|W_{g-1})$ represents the likelihood that a particular synaptic cluster, $\bar{s}_{g,c}$, be synthesized as a part of the network architecture in generation $g$ given the synaptic strength in generation $g-1$.  For example, in a deep convolutional neural network, the synaptic cluster $c$ can be any subset of synapses such as a kernel or a set of kernels within the deep neural network.  The probability $P(s_{g,i}|w_{g-1,i})$ represents the likelihood of existence of synapse $i$ within the cluster $c$ in generation $g$ given its synaptic strength in generation $g-1$.  As such, the proposed synaptic probability model not only promotes the persistence of strong synaptic connectivity in offspring deep neural networks over successive generations, but also promotes the persistence of strong synaptic clusters in offspring deep neural networks over successive generations.
%

\textbf{Cluster-driven Evolutionary Synthesis.}  In the seminal paper on evolutionary deep intelligence by Shafiee~{\it et al.}~\cite{EvoNet1}, the synthesis probability $P(\mathcal{H}_g)$ is composed of the synaptic probability $P(S_g|W_{g-1})$, which mimic heredity, and environmental factor model $\mathcal{F}(\mathcal{E})$ which mimic natural selection by introducing quantitative environmental conditions that offspring networks must adapt to:
\vspace{-0.05 cm}
\begin{align}
P(\mathcal{H}_g) = \mathcal{F}(\mathcal{E}) \cdot P(S_g|W_{g-1})
\label{eq:EvoNet_EnvFactor}
\end{align}
In this study, \eqref{eq:EvoNet_EnvFactor} is reformulated in a more general way to enable the incorporation of different quantitative environmental factors over both the synthesis of synaptic clusters as well as each synapse:
\vspace{-0.55 cm}
\begin{align}
\vspace{-0.15 cm}
\label{eq:NewEvoNet_EnvFactor}
P(\mathcal{H}_g) =
\prod_{c \in C} \Big [\mathcal{F}_c (\mathcal{E}) P\big(\bar{s}_{g,c}|W_{g-1}\big) \cdot \prod_{i \in c} \mathcal{F}_s(\mathcal{E}) P(s_{g,i}|w_{g-1,i}) \Big]
\vspace{-0.45 cm}
\end{align}
where $\mathcal{F}_c (\cdot) $ and $\mathcal{F}_s(\cdot) $  represents environmental factors enforced at the cluster and synapse levels, respectively.

\textbf{Realization of Cluster-driven Genetic Encoding.}
In this study, a simple realization of the proposed cluster-driven genetic encoding scheme is presented to demonstrate the benefits of the proposed scheme.  Here, since we wish to promote the persistence of strong synaptic clusters in offspring deep neural networks over successive generations, the probability of the synthesis of a particular cluster of synapses, $\bar{s}_{g,c}$ is modeled as
\vspace{-0.3 cm}
\begin{align}
P\big(\bar{s}_{g,c} =1|W_{g-1}\big) = \exp \Big( \frac{\sum_{i \in c}\lfloor{\omega_{g-1,i}}\rfloor}{Z} - 1 \Big)
\label{eq:CLGE}
\end{align}
where $\lfloor{\cdot}\rfloor$ encodes the truncation of a synaptic weight and $Z$ is a normalization factor to make~\eqref{eq:CLGE} a probability distribution,  $P\big(\bar{s}_{g,c}|W_{g-1}\big)  \in [0,1]$.  The truncation of synaptic weights in the model reduces the influence of very weak synapses within a synaptic cluster on the genetic encoding process.  The probability of a particular synapse, $s_{g,i}$, within synaptic cluster $c$, denoted by $P(s_{g,i} = 1|w_{g-1,i})$ can be expressed as:
\vspace{-0.4 cm}
\begin{align}
P(s_{g,i} = 1|w_{g-1,i})  = \exp\Big( \frac{ \omega_{g-1,i}}{z} - 1 \Big)
\end{align}
where $z$ is a layer-wise normalization constant. By incorporating both of the aforementioned probabilities in the proposed scheme, the relationships amongst synapses as well as their individual synaptic strengths are taken into consideration in the genetic encoding process.
\vspace{-0.4 cm}
\section{Experimental Results}
\vspace{-0.4 cm}

 Evolutionary synthesis of deep neural networks across several generations is performed using the proposed genetic encoding scheme, and their network architectures and accuracies are investigated using three benchmark datasets: MNIST~\cite{MNIST}, STL-10~\cite{STL10} and CIFAR10~\cite{CIFAR10}.
The LeNet-5 architecture~\cite{MNIST} is selected as the network architecture of the original, first generation ancestor network for MNIST and STL-10, while the AlexNet architecture~\cite{krizhevsky2012imagenet} is utilized for the ancestor network for CIFAR10, with the first layer modified to utilize $5 \times 5 \times 3$ kernels instead of $11 \times 11 \times 3$ kernels given the smaller image size in CIFAR10.
The environmental factor model being imposed at different generations in this study is designed to form deep neural networks with progressively more efficient network architectures than its ancestor networks while maintaining modeling accuracy.
More specifically, $\mathcal{F}_c(\cdot)$ and $\mathcal{F}_s(\cdot)$ is formulated in this study such that an offspring deep neural network should not have more than 80\% of the total number of synapses in its direct ancestor network.
Furthermore, in this study, each kernel in the deep neural network is considered as a synaptic cluster in the synapse probability model.  In other words, the probability of the synthesis of a particular synaptic cluster (i.e, $P(\bar{s}_{g,c}|W_{g-1})$) is modeled as the truncated summation of the weights within a kernel.
%
%

\textbf{Results \& Discussion}. In this study, offspring deep neural networks were synthesized in successive generations until the accuracy of the offspring network exceeded 3\%, so that we can better study the changes in architectural efficiency in the descendant networks over multiple generations.  Table~\ref{Tab:Mnist-STL10Res} shows the architectural efficiency (defined in this study as the total number of synapses of the original, first-generation ancestor network divided by that of the current synthesized network) versus the modeling accuracy at several generations for three datasets. As observed in Table~\ref{Tab:Mnist-STL10Res}, the descendant network at the 13th generation for MNIST was a staggering $\sim$125-fold more efficient than the original, first-generation ancestor network without exhibiting a significant drop in the test accuracy ($\sim$1.7\% drop).  This trend was consistent with that observed with the STL-10 results, where the descendant network at the 10th generation was $\sim$56-fold more efficient than the original, first-generation ancestor network without a significant drop in test accuracy ($\sim$1.2\% drop). It also worth noting that since the training dataset of the STL-10 dataset is relatively small, the descendant networks at generations 2 to 8 actually achieved higher test accuracies when compared to the original, first-generation ancestor network, which illustrates the generalizability of the descendant networks compared to the original ancestor network as the descendant networks had fewer parameters to train.  Finally, for the case of CIFAR10 where a different network architecture was used (AlexNet), the descendant network at the 6th generation network was $\sim$14.4-fold more efficient than the original ancestor network with $\sim$2\% drop in test accuracy, thus demonstrating the applicability of the proposed scheme for different network architectures.
\vspace{-0.2 cm}
\begin{table}
	\vspace{-0.45 cm}
	\caption{Architectural efficiency vs. test accuracy for different generations of synthesized networks. ``Gen.'', ``A-E'' and ``ACC.'' denote generation, architectural efficiency, and accuracy, respectively.  }
	\vspace{-0.15cm}
	\label{Tab:Mnist-STL10Res}
\begin{center}
\begin{minipage}{.3\textwidth}
\setlength\tabcolsep{0.15 cm}
	\begin{tabular}{|c||c|c|}
	
				\multicolumn{3}{c}{\textbf{\footnotesize MNIST}} \\\hline
			\footnotesize Gen.    & \footnotesize A-E & \footnotesize ACC.  \\ \hline \hline
			1   &1.00X         &0.9947 \\
			5    & 5.20X  & 0.9941 \\
			7  & 12.09X    &0.9928\\
			9   &  32.23X  & 0.9884\\
			11 &   62.74X    &0.9849 \\
			13   &   125.09X & 0.9775\\\hline
		\end{tabular}
\end{minipage}\hfill
\begin{minipage}{.3\textwidth}
 \setlength\tabcolsep{0.15 cm}
\begin{tabular}{|c||c|c|}
			\multicolumn{3}{c}{\textbf{\footnotesize STL-10}} \\\hline
			Gen.   & \footnotesize A-E &\footnotesize ACC.  \\ \hline \hline
			1  & 1.00X    & 0.5774\\
			3 &  2.37X   &0.5933\\
			5    &  5.81X   &  0.6039   \\
			7   &  14.99X   &0.6051\\
			9  &   38.22X  &0.5744\\
			10   &   56.27X  &0.5658\\\hline
		\end{tabular}
\end{minipage}\hfill
\begin{minipage}{.3\textwidth}
	\setlength\tabcolsep{0.15 cm}
	\begin{tabular}{|c||c|c|}
		\multicolumn{3}{c}{\textbf{\footnotesize CIFAR10}} \\\hline
		Gen.    &\footnotesize A-E &\footnotesize ACC.  \\ \hline \hline
		1    &  1.00X   & 0.8669\\
		2    &   1.64X  & 0.8814\\
		3   &  2.82X   &0.8766 \\
		4  &   5.06X  & 0.8688\\
		5   &   8.39X  & 0.8588\\
		6   &   14.39X  & 0.8459\\\hline

	\end{tabular}
\end{minipage}\hfill
\vspace{- 0.25 cm}
\end{center}
\end{table}

\begin{table}[]
	\vspace{-0.05cm}
	\setlength\tabcolsep{0.1 cm}
	\begin{center}
		\footnotesize
		\caption{Cluster efficiency of the convolutional layers (layers 1-3) and fully connected layer (layer 4) at first and the last reported generations of deep neural networks for MNIST and STL-10. Columns `E' show overall cluster efficiency for synthesized deep neural networks.}
		\vspace{-0.30 cm}
		\label{Tab:numKernelRes}
		\setlength\tabcolsep{0.1 cm}
		\begin{tabular}{|l|c|cccc|c||l|c|cccc|c|}
			
			\multicolumn{6}{c}{\textbf{MNIST}} & 	\multicolumn{6}{c}{\textbf{STL-10}}\\\hline 
			Gen. & ACC.& Layer 1 & Layer 2&Layer 3&Layer 4&  E & Gen. & ACC.& Layer 1 & Layer 2&Layer 3&Layer 4 & E \\\hline 
			1({\tiny baseline})&0.9947&1X&1X&1X&1X&  1X	&1({\tiny baseline})& 0.5774&1X&1X&1X&1X&1X\\
			13& 0.9775&3.6X&12.96X&8.83X&13.47X &9.71X & 10&0.5658&3.31X&8.46X&5.49X&6.56X & 5.96X\\\hline
		\end{tabular}
	\end{center}
	\vspace{-0.35 cm}
\end{table}

\begin{table}[!t]
	\vspace{-0.05 cm}
	\setlength\tabcolsep{0.1 cm}
	\begin{center}
		\footnotesize
		\caption{Cluster efficiency of the convolutional layers (layers 1-5) and fully connected layers (layers 6-7) at first and the last reported generations of deep neural networks for CIFAR10.}
		\label{Tab:Cifar}
		\begin{tabular}{|l|c|ccccccc|c|}
			\hline
			Gen. & ACC.&  Layer 1 & Layer 2&Layer 3&Layer 4 &Layer 5&Layer 6&Layer 7 & E \\\hline
			1 ({\tiny baseline})&0.8669&1X & 1X &1X & 1X &1X &1X &1X &1X \\
			6&0.8459&2.01X&3.10X&3.50&3.40X& 3.29X&2.97X&1.51X&2.82X \\\hline
		\end{tabular}
	\end{center}
	\vspace{-0.35 cm}
\end{table}
\textbf{Embedded GPU Ramifications}. Table~\ref{Tab:numKernelRes} and~\ref{Tab:Cifar} shows the cluster efficiency per layer of the synthesized deep neural networks in the last generations, where cluster efficiency is defined in this study as the total number of kernels in a layer of the original, first-generation ancestor network divided by that of the current synthesized network. It can be observed that for MNIST, the cluster efficiency of last-generation descendant network is $\sim$9.7X, which may result in a near 9.7-fold potential speed-up in running time on embedded GPUs by reducing the number of arithmetic operations by $\sim$9.7-fold compared to the first-generation ancestor network, though computational overhead in other layers such as ReLU may lead to a reduction in actual speed-up.  The potential speed-up from the last-generation descendant network for STL-10 is lower compared to MNIST dataset, with the reported cluster efficiency in last-generation descendant network  $\sim$6X.  Finally, the cluster efficiency for the last generation descendant network for CIFAR10 is $\sim$2.8X, as shown in Table~\ref{Tab:Cifar}. These results demonstrate that not only can the proposed genetic encoding scheme promotes the synthesis of deep neural networks that are highly efficient yet maintains modeling accuracy, but also promotes the formation of highly sparse synaptic clusters that make them highly tailored for devices designed for highly parallel computations such as embedded GPUs.

%
\vspace{-0.05 cm}
\subsubsection*{Acknowledgments}
\vspace{-0.35 cm}
This research has been supported by Canada Research Chairs programs, Natural Sciences and Engineering Research Council of Canada (NSERC), and the Ministry of Research and Innovation of Ontario.  The authors also thank Nvidia for the GPU hardware used in this study through the Nvidia Hardware Grant Program.






\end{document}